\documentclass[runningheads]{llncs}
\usepackage{graphicx}

\usepackage{tikz}
\usepackage{comment}
\usepackage{amsmath,amssymb} 
\usepackage{color} 

\usepackage[accsupp]{axessibility}  
\usepackage{subcaption,siunitx}
\usepackage{wrapfig}

\begin{document}
\pagestyle{headings}
\mainmatter
\def\ECCVSubNumber{2319}  

\title{MeshMAE: Masked Autoencoders for 3D Mesh Data Analysis} 

\titlerunning{MeshMAE: Masked Autoencoders for 3D Mesh Data Analysis}
%
\author{Yaqian Liang \inst{1} \index{van Author, First E.}\thanks{This work was done during Y. Liang’s internship at JD Explore Academy.} \and
Shanshan Zhao\inst{2} \and
Baosheng Yu\inst{3} \and
Jing Zhang\inst{3}  \and \\
 Fazhi He\inst{1} 
 }
\authorrunning{Y. Liang et al.}
%
\institute{School of Computer Science, Wuhan University,  China\\
\and JD Explore Academy,  China\\
\and
School of Computer Science, The University of Sydney, Australia\\
\email{\{yqliang, fzhe\}@whu.edu.cn}, \email{\{sshan.zhao00\}@gmail.com}\\
\email{\{baosheng.yu, jing.zhang1\}@sydney.edu.au}}


\maketitle

\begin{abstract}

Recently, self-supervised pre-training has advanced Vision Transformers on various tasks \textit{w.r.t.} different data modalities, \textit{e.g.}, image and 3D point cloud data. In this paper, we explore this learning paradigm for 3D mesh data analysis based on Transformers. Since applying Transformer architectures to new modalities is usually non-trivial, we first adapt Vision Transformer to 3D mesh data processing, \textit{i.e.,} Mesh Transformer. In specific, we divide a mesh into several non-overlapping local patches with each containing the same number of faces and use the 3D position of each patch's center point to form positional embeddings. Inspired by MAE, we explore how pre-training on 3D mesh data with the Transformer-based structure benefits downstream 3D mesh analysis tasks. We first randomly mask some patches of the mesh and feed the corrupted mesh into Mesh Transformers. Then, through reconstructing the information of masked patches, the network is capable of learning discriminative representations for mesh data. Therefore, we name our method MeshMAE, which can yield state-of-the-art  or comparable performance on mesh analysis tasks, \textit{i.e.,} classification and segmentation. In addition, we also conduct comprehensive ablation studies to show the effectiveness of key designs in our method.

\end{abstract}
\section{Introduction}

In recent years, Transformers \cite{vaswani2017attention} have been the technological dominant architecture in NLP community \cite{brown2020language,joshi2020spanbert,devlin2018bert}. With autoregressive language
modeling~\cite{radford2018improving,radford2019language} or masked autoencoding~\cite{devlin2018bert}, Transformer architectures can be trained on a very large unlabeled corpus of text. In computer vision, to pre-train a generalizable vision model, many strategies have been intensively studied, such as contrastive learning~\cite{hjelm2018learning}, egomotion prediction~\cite{agrawal2015learning}, and geometric transformation recognition~\cite{gidaris2018unsupervised}. Recently, inspired by the success of the masked autoencoding strategy in NLP, some works~\cite{bao2021beit,yu2021point} also apply such a strategy to images or point clouds   by reconstructing the masked tokens. Instead of predicting the tokens, a latest work, MAE~\cite{he2022masked}, proposes to directly reconstruct raw pixels with a very impressive performance achieved, which provides a new self-supervised pre-training paradigm. Inspired by this, in this paper, we further explore the masked autoencoding strategy on 3D mesh data, whose structure is different from either 2D images or 3D point clouds.

As an effective 3D representation, 3D mesh has been widely exploited in computer graphics, such as 3D rendering, model reconstruction, and animation ~\cite{hu2020jittor,guan2021bilevel,LuanWZWZQ21}. Along with the development of deep learning, remarkable achievements have been made in various mesh analysis tasks by adopting deep neural networks, such as 3D mesh classification \cite{feng2019meshnet},  segmentation \cite{hu2021subdivision}, and 3D human  reconstruction \cite{lin2021end}. 
In comparison with a image, the 3D mesh is composed of vertices and faces, which have not the specific order. While  the connection between vertices makes 3D mesh not completely discrete data as point cloud. In light of the distinct structure of mesh data,  there are some key challenges that need to be solved before applying the masked autoencoding strategy to the 3D mesh.

In this paper, we start from adapting the vanilla Vision Transformer~\cite{dosovitskiy2020image} to process mesh data, which remains unexplored. As there are lots of faces in a mesh, \textit{e.g.,} from $10^3$ to $10^5$ in the original ModelNet40 (or ShapeNet), it is almost infeasible to apply the self-attention mechanism in Transformers over all faces. Inspired by~\cite{dosovitskiy2020image}, we split the original mesh into a set of non-overlapping mesh patches, which are regarded as the basic unit processed by the Transformer block. In addition, we calculate the positional embeddings in a naive manner since 3D mesh data inherently contains positional information. Specifically, the 3D coordinate of the center point of each mesh patch is used as the positional embedding. Finally, built upon the vanilla Vision Transformers, our Mesh Transformer processes these mesh embeddings via the multi-head self-attention layer and feedforward network sequentially. Basically, exploiting these operations, we can apply the Transformer architecture for 3D mesh data, \textit{i.e.,} Mesh Transformer.

Inspired by MAE \cite{he2022masked}, we  wonder whether performing masked autoencoding on the unlabeled 3D mesh data could also promote the  ability of the network. 
To arrive at this, we are required to design an efficient reconstruction objective first. Unlike MAE~\cite{he2022masked}, which reconstructs the raw pixels naturally, we choose to recover the geometric information of the masked patches. In detail, we achieve the reconstruction at two levels, \textit{i.e.,} point-level and face-level. For point-level, we aim to predict the position of all points belonging to the masked patch, while for face-level, we regress the features for each face.
Taking advantage of the two reconstruction objectives, the model can learn discriminative mesh representation in a self-supervised manner from large-scale unlabeled 3D mesh data.

On the basis of the above analysis, we present MeshMAE, a masked autoencoder network for Transformer-based 3D mesh pre-training. To  show the effectiveness of the proposed pre-text task, we  visualize the reconstructed meshes. We observe that our MeshMAE model correctly predicts the shape of masked mesh patches and infers diverse, holistic reconstructions through our decoder. We conduct the pre-training on ModelNet40 and ShapeNet respectively and conduct the fine-tuning and linear probing experiments  for mesh classification. In addition, we also finetune the pre-trained network for the part segmentation task. The experimental results and comprehensive ablations demonstrate the effectiveness of our method. The main contributions of our method are as follows:

\begin{itemize}
    
    \item We explore adapting the masked-autoencoding-based pre-training strategy to 3D mesh analysis. To achieve this, we design a feasible Transformer-based network, Mesh Transformer, and reconstruction objectives in light of the distinctive structure of 3D mesh data.

    \item Our method achieves new state-of-the-art or comparable performance on the mesh classification and segmentation tasks. We also conduct intensive ablations to better understand the proposed MeshMAE.
    
\end{itemize}


\section{Related works}

\subsection{Transformer}
 
Transformers  \cite{vaswani2017attention} were first introduced as an attention-based framework in NLP and currently have become the dominant framework in NLP \cite{devlin2018bert,radford2019language} due to its salient benefits, including massively parallel computing, long-distance characteristics, and minimal inductive bias. Along with its remarkable success in NLP,  the trend towards leveraging Transformer architecture into the computer vision community has also emerged. The Vision Transformer (ViT) \cite{dosovitskiy2020image} makes a significant attempt to exploit the self-attention mechanism,
and has obtained many state-of-the-art (SOTA) records. So far, Transformers have already been extensively studied in various computer tasks, especially 2D image analysis tasks, such as image classification \cite{chen2021crossvit,dosovitskiy2020image,touvron2021training,xu2021vitae,zhang2022vitaev2}, object detection \cite{zhu2020deformable,carion2020end,dai2021up,misra2021end},  semantic segmentation \cite{wang2021max,zheng2021rethinking,cheng2021per,ding2021looking}, pose estimation \cite{lin2021end,huang2020hand,li2021mhformer},  and depth estimation \cite{li2021revisiting,yang2021transformer}.
However, the application of ViT in  3D data still remains limited. 

Almost simultaneously, \cite{guo2021pct,yu2021pointr,zhao2021point,trappolini2021shape} proposes to extend Transformer architectures into the disordered and discrete 3D point cloud. 
Among them, \cite{guo2021pct} proposes an offset attention (OA) module to sharpen the attention weights and reduce the influence of noise. 
To save the computational costs, \cite{zhao2021point} proposes a  local vector self-attention mechanism to construct a point Transformer layer. 
In comparison, to facilitate Transformer to better leverage the inductive bias  of point clouds, \cite{yu2021pointr} devises a geometry-aware block to model the local geometric relationships.
Nevertheless, those prior efforts all involve more or less inductive biases, making them out of  line with standard Transformers. As for meshes, to the best of our knowledge, there are no existing papers that directly apply Transformers to process the irregular mesh data. Instead, some works only use images as the input of the Transformer and output mesh data. For example, MeshGraphormer \cite{lin2021mesh} combines the graph convolutions and self-attentions to model both local and global interactions,  
METRO \cite{lin2021end} proposes to utilize Transformer to learn the correlation among long-range joints and vertices in images. 
PolyGen \cite{nash2020polygen} proposes to generate the vertices and faces in sequence with Transformer. 
In this work, we would explore how to apply the standard Transformer to 3D mesh data.

\subsection{Self-supervised Learning} 

Self-supervised learning could learn the meaningful feature representations via pretext tasks that do not require extra annotations. For example, contrastive  learning learns feature representations by increasing intra-class distance and decreasing extra-class distance \cite{wang2015unsupervised,pathak2017learning}. In comparison, autoencoding pursues a conceptually different direction by mapping an input to a latent representation in an encoder and reconstructing the input using a decoder. In NLP, BERT \cite{devlin2018bert} first attempts to pre-train bidirectional representations from the unlabeled text in a self-supervised scheme. Since then, the pretext of Masked Language Modeling (MLM) has arisen  significant interest in NLP \cite{brown2020language,radford2018improving,radford2019language}. Motivated by the success of BERT in NLP, there are many attempts in the  computer vision area \cite{bao2021beit,chen2020generative,conneau2019cross,joshi2020spanbert,liu2019roberta,trinh2019selfie}.  Among them, iGPT \cite{chen2020generative} proposes to operate on sequences of pixels and predict the unknown pixels, while ViT \cite{dosovitskiy2020image} proposes to reshape the images into patches, and then adapts the standard Transformer to process images. Recently,  BEiT \cite{bao2021beit}  proposes a pretext task that tokenizes the input images into discrete visual tokens firstly and then recovers the masked discrete tokens, while MAE \cite{he2022masked}  encourages the model to reconstruct those missing pixels directly without the tokenization.
For 3D data analysis, PointBERT \cite{yu2021point} generalizes the concept of BERT  into 3D point clouds by devising a masked point reconstruction task to pre-train the Transformers. Variation autoencoder (VAE) has been applied to implement 3D mesh deformation and pose transfer \cite{cosmo2020limp,aumentado2019geometric}. Our work is greatly inspired by MAE and PointBERT.  
However, the distinctive characteristics of 3D meshes hinder the straightforward use of BERT on 3D mesh analysis. In this paper, we aim to explore how to adapt mask autoencoding pre-training to 3D meshes with minimal modification.

\subsection{Mesh Analysis}

The polygon mesh is an effective 3D  representation, which can depict the geometric context of a 3D object preciously. However, the vertices in 3D meshes do not have the same number of adjacent vertices, leading to the general convolution and pooling operations cannot be applied to the mesh data directly. 
Accompanied by the development of deep learning methods on 3D data, how to adapt the neural network to 3D mesh processing is always the highlight.  
Initially, Masci~\emph{et al.}~\cite{2015Learning} makes the first effort to generalize convolutional neural networks to 3D mesh based on localized frequency analysis.  More recently, researchers extensively study how to implement feature aggregation and downsampling operations on irregular structures. For example, MeshCNN \cite{hanocka2019meshcnn:} explores classical methods of Graphics,  which defines an ordering invariant convolution operation for edges and utilizes the edge collapse operation to implement pooling. To overcome the problem that the unfixed number of vertex neighborhoods in the mesh, some methods introduce  graph neural networks (GNNs) \cite{milano2020primal,verma2018feastnet:,monti2018dual,saleh2022bending} to implement the convolution on vertices, where the feature aggregation is conducted on the vertices and their 1-ring neighbors.  
While each face must have three adjacent faces in the manifold mesh, this kind of stable structure makes feature aggregation of faces easier.
MeshNet learns the spatial and structural features of a face by aggregating its three adjacent faces. SubdivNet \cite{hu2021subdivision} processes the meshes to obtain the fine-to-coarse hierarchical structure, which could be processed by the convolution-like operation.   
Actually, the attention-based Transformer does not require convolution or pooling operation and would not be affected by the data structure. Therefore, it may be more suitable for irregular mesh data.

\section{Method}

\begin{figure*}[t]
    \centering
    \includegraphics[width=0.85\linewidth]{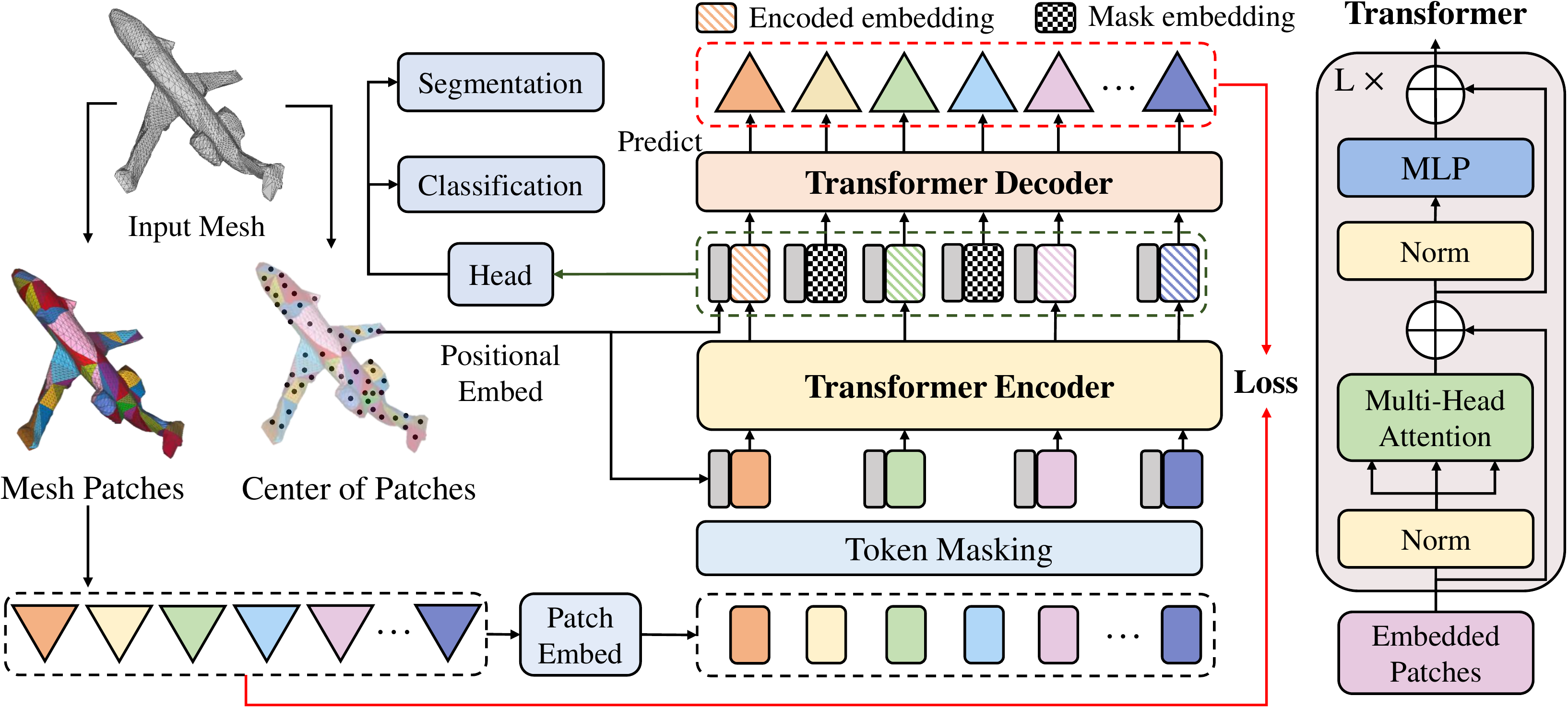}
    \caption{The overall architecture of the proposed MeshMAE framework. The input mesh is divided into several non-overlapping patches, which would be embedded by the MLP. During pre-training, patch embeddings will be randomly masked, and only the visible embeddings will be fed into the Transformers. Then, the mask embeddings are introduced and sent to the decoder together with the combination of encoded embeddings. The targets of the decoder are to reconstruct the vertices and face features of masked patches.  After pre-training, the decoder is discarded and the encoder is applied to downstream tasks.}
    \label{fig:backbone}
\end{figure*}

In this paper, we aim to extend the Vision Transformer into mesh analysis, Mesh Transformer, and apply the masked-autoencoding-based self-supervised pre-training strategy to Mesh Transformer. Here, we introduce the details of our Transformer-based mesh analysis framework, Mesh Transformer, and the MAE-based pre-training strategy. Figure \ref{fig:backbone} shows the framework of MeshMAE. 

\subsection{Mesh Transformer}

A triangle mesh $M = (V, F)$ is defined by vertices  and triangular faces. To apply Transformer on mesh, we need to address patch split, patch embedding, and positional embedding, which are introduced in the following parts.

\subsubsection{Mesh Patch Split.}

In comparison with 3D point cloud data containing a set of discrete points, the faces of 3D mesh provide the connection relationship between vertices. As a result, we can use the geometric information of each face to represent the feature. In detail, as SubdivNet \cite{hu2021subdivision} does, for face $f_i$, we define its feature as a 10-dimensional vector, consisting of the face area (1-dim),  three interior angles of the triangle (3-dim), face normal (3-dim), and  three inner products between the face normal and three vertex normals (3-dim).

\begin{figure}[!ht]
    \centering
    \includegraphics[width=0.9\linewidth,height=1.5in]{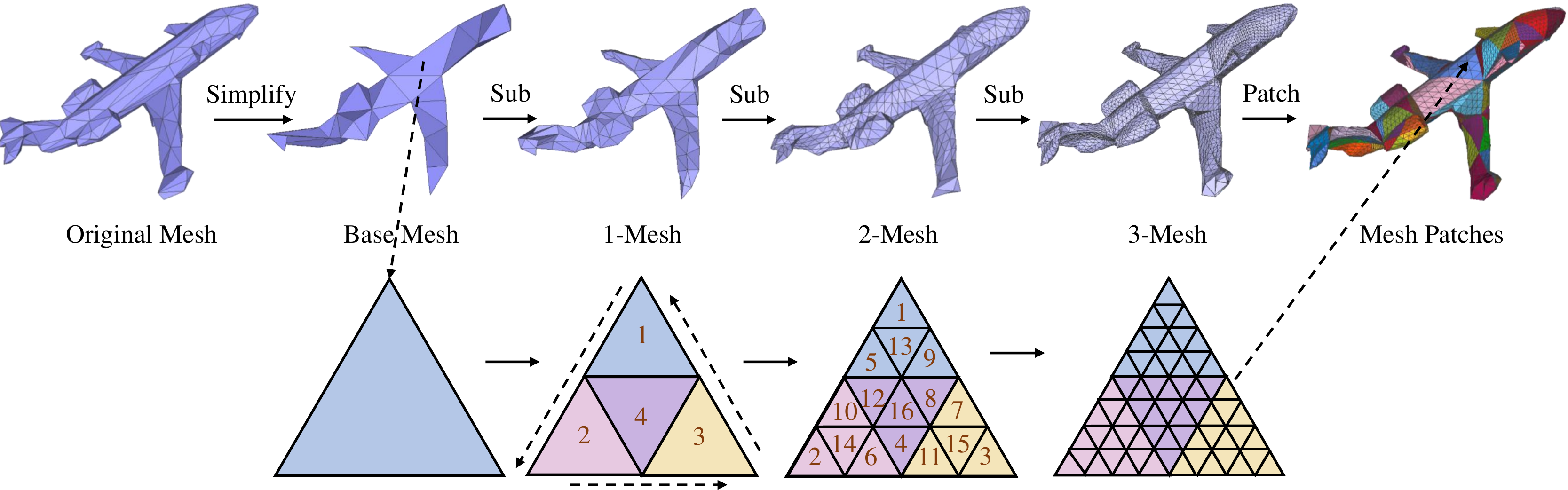}
    \caption{The process of remeshing operation. The input mesh is simplified first, and a bijection is constructed between the original mesh and the base mesh. Then the base mesh is subdivided 3 times and new vertices are projected back onto the input.}
    \label{fig:patch1}
\end{figure}

The self-attention-based architectures of Transformers ease the pains of designing especial feature aggregation operations for 3D meshes. However, if we apply the self-attention mechanism over all faces, the huge computational cost of quadratic complexity would be unbearable. Therefore, before applying Transformers to mesh, we first group the faces of a mesh into a set of non-overlapping patches. However, unlike image data, which is regular and could be divided into a grid of square patches, mesh data is irregular and the faces are usually unordered, which makes the patch split challenging.  To address this issue, we propose to first `remesh' the original mesh to make the structure regular and hierarchical. Specifically, we adopt MAPS algorithm \cite{lee1998maps} to implement the remeshing operation, which simplifies the mesh to a base mesh with $N$ faces ($96\leq N\leq 256$ in our experiments). After the remeshing process, the base mesh, \textit{simplified mesh}, is coarser than the original mesh and is incapable of representing the shape accurately. Therefore, we further subdivide all faces in the base mesh $t$ times in a 1-to-4 manner and get a refined mesh called $t-$mesh. We can divide $t-$mesh into non-overlapping patches by grouping the faces corresponding to the same face in the base mesh into a patch. In practice,  we subdivide 3 times and thus each patch contains 64 faces. The process is illustrated in Figure~\ref{fig:patch1}.

\subsubsection{Transformer Backbone.}
In this paper, we adopt the standard Transformer as the backbone network, consisting of a series of multi-headed self-attention layers and feedforward network (FFN) blocks. 

To represent each patch, for image data, we can concatenate the  RGB values of raw pixels located in the same patch in a natural order. For mesh data, thanks to remeshing operation which subdivides the face orderly, we can also use the concatenation of the 10-dim feature vectors of all faces belonging to the same patch as the patch's feature. As shown in Figure~\ref{fig:patch1}, we can find that the face in the base mesh is always divided in a fixed order starting from a selected vertex. Although the selected vertex might impact the order, we experimentally find that the impact is very slight (Sec. \ref{faceorder}). Therefore, we can concatenate the faces' features according to the order generated by the remeshing process as the patch's representation (as shown in Figure~\ref{fig:order}). Then, we adopt an MLP to project the feature vector of those patches into representations ${\{e_i\}}^g_{i=1}$, where $g$ denotes the number of patches. The representations are thus regarded as the input into the Transformer. The input features only describe the shape information of faces. Generally, Transformer-based methods utilize the positional embedding to provide the position information of patches.  In comparison against natural language or image  where the serial number is taken as the position, the mesh data contains 3D space position of each face. 
In this paper, we adopt the center 3D coordinates of faces to compute the positional embeddings, which would be more suitable for  the geometric data and unordered patches. In practical, we first calculate the center point coordinates ${\{c_i\}}^g_{i=1}$ of each patch, and then apply an MLP to   ${\{c_i\}}^g_{i=1}$  to obtain the positional embedding ${\{p_i\}}^g_{i=1}$ for patches.

Formally, we define the input embeddings  ${X = \{x_i\}}^g_{i=1}$ as  the combination of patch embeddings ${E =\{e_i\}}^g_{i=1}$ and positional embeddings $P =\{p_i\}^g_{i=1}$.  
In this way, the overall input sequence is defined as $H^0 = \{x_1, x_2, ..., x_g\}$. There are $L$ layers of Transformer block in the encoder network, and the output of the last layer $H^L=\{h^L_1, ..., h^L_g\}$ represents the encoded representation of the input patches.

\subsection{Mesh Pre-training Task}
Motivated by BERT \cite{devlin2018bert} and MAE \cite{he2022masked}, we study the masked modeling strategy for mesh representation learning based on the introduced Mesh Transformer. Specifically, we devise a masked mesh modeling task that aims at reconstructing the geometric structure of the masked patches from partially visible sub-meshes. Our method is based on autoencoders, where an encoder is utilized to map the visible sub-meshes into latent representations and the decoder reconstructs the geometric structure from the latent representations. By modeling the masked parts, the model attains the geometric understanding of the mesh. Below, we provide more details.

\subsubsection{Encoder and Decoder.}

In the pre-training task, the encoder and decoder networks are both composed of several Transformer blocks. We set the encoder, \textit{i.e.,} our Mesh Transformer, as 12 layers and a lightweight decoder with 6 layers. In our method, according to a pre-defined masking ratio, some patches of the input mesh are masked, and the remaining visible patches are fed into the encoder. Before  feeding into the decoder, we utilize a shared mask embedding to replace all the masked embeddings, which indicates the presence of the missing patches that need to be predicted. Then, the input to the decoder is composed of encoded visible embeddings and mask embeddings. Here,  we add the positional embeddings to all embeddings once again, which provide the location information for both masked and visible patches. It is noted that the decoder is only used during pre-training to perform mesh reconstruction tasks, while only the encoder is used in the downstream tasks. 

\subsubsection{Masked Sequence Generation.} 

The complete mesh embeddings are denoted by $E = {\{e_1, ..., e_g\}}$ and their indices  are denoted by $I = {\{1, ..., g\}}$. Following MAE, we first randomly mask a subset of patches, where the indices $I_m$ of masked embeddings are sampled from $I$ randomly with the ratio $r$. In this way, we denote the masked embeddings as $E_m = E[I_m]$ and the unmasked embeddings as $E_{um} = E[I-I_m]$.  Next, we replace the  masked embeddings $E_m$ with a shared learnable mask embedding $E_{mask}$ while keeping their positional embedding unchanged. Finally, the corrupted mesh embeddings $E_c = E_{um} \cup \{E_{mask} + p_i: i \in I_m\}$  are fed into the encoder.
When the masking ratio is  high, the redundancy can be eliminated largely and the masked parts would not be solved easily by extrapolation from visible neighboring patches. Therefore, the reconstruction task is relatively challenging.

\subsubsection{Reconstruction Targets.}

The targets of the pre-training task are also crucial. In NLP, BERT \cite{devlin2018bert} proposes the pretext of Masked Language Modeling (MLM), which first masks the input randomly and then recovers a sequence of input tokens. Following it, there are also similar works on images, \textit{e.g.}, BEiT \cite{bao2021beit},  and on point clouds, \textit{e.g.,} PointBert \cite{yu2021point}, both of which train the models by recovering the tokens learned via dVAE \cite{rolfe2016discrete}.  However, these kinds of dVAE tokenizer reconstruction require one more pre-training stage. Recently, MAE \cite{he2022masked} skips the dVAE training process and proposes to reconstruct the pixel values directly, which is simpler and saves much computation overhead.  

Inspired by  MAE \cite{he2022masked}, our method directly recovers the input patches.  Specifically, we define a reconstruction target as the shape of the masked patches. As shown in Figure \ref{fig:patch1}, there are only 45 unique vertices in the 64 faces of a mesh patch. To recover the shape of patches, these 45 vertices are required to be located in the corresponding ground truth positions. Therefore, we propose to predict the 3D relative coordinate $(x, y, z)$ of them directly (relative to the center point of the patch), and the output of the decoder can be written as $P_r = \{{p_{r_i}\}}^{45}_{i=1}$. 
During the training phase, we adopt the $l_2$-form Chamfer distance  to calculate the reconstruction loss  between the relative coordinates of predicted vertices and that of ground truth vertices, which is shown as follows:
\begin{equation}
    \mathcal{L}_{CD}(P_r, G_r) = \frac{1}{|P_r|}\sum_{p\in P_r}\underset{g \in G_r}{min}||p-g|| + \frac{1}{|G_r|}\sum_{g\in G_r}\underset{p \in P_r}{min}||g-p||, 
\end{equation}
where $G_r$ denotes the relative coordinates of the 45  ground truth vertices in a patch.

In the real mesh, the vertices are connected with edges to compose the faces, which are the important unit of the mesh data.  However, due to the disorder of the vertices, 
it is intractable to recover the faces' information if we only predict the vertices' position, which might degrade the network's capability of recovering the geometric structure.
In order to enhance the restraint and predict the local details, we propose to additionally predict the face-wise features, \textit{i.e.,} the input representation of the face. 
To achieve this, we add another linear layer behind the decoder to predict all faces' features for each patch. We denote the predicted features by $J\in\mathcal{R}^{64\times10}$. Here, we organize the faces' features in the order generated by the remeshing operation. 
We adopt the face-wise MSE loss $\mathcal{L}_{MSE}$ to evaluate the reconstruction effect of features. The overall optimization object is combined by both $\mathcal{L}_{CD}$ and $\mathcal{L}_{MSE}$, as follows,
\begin{equation}
    \mathcal{L} = \mathcal{L}_{MSE} + \lambda  \cdot \mathcal{L}_{CD},
\end{equation}
where $\lambda$ is the loss weight.

\begin{figure*}[ht]
    \centering
    \includegraphics[width=0.85\linewidth]{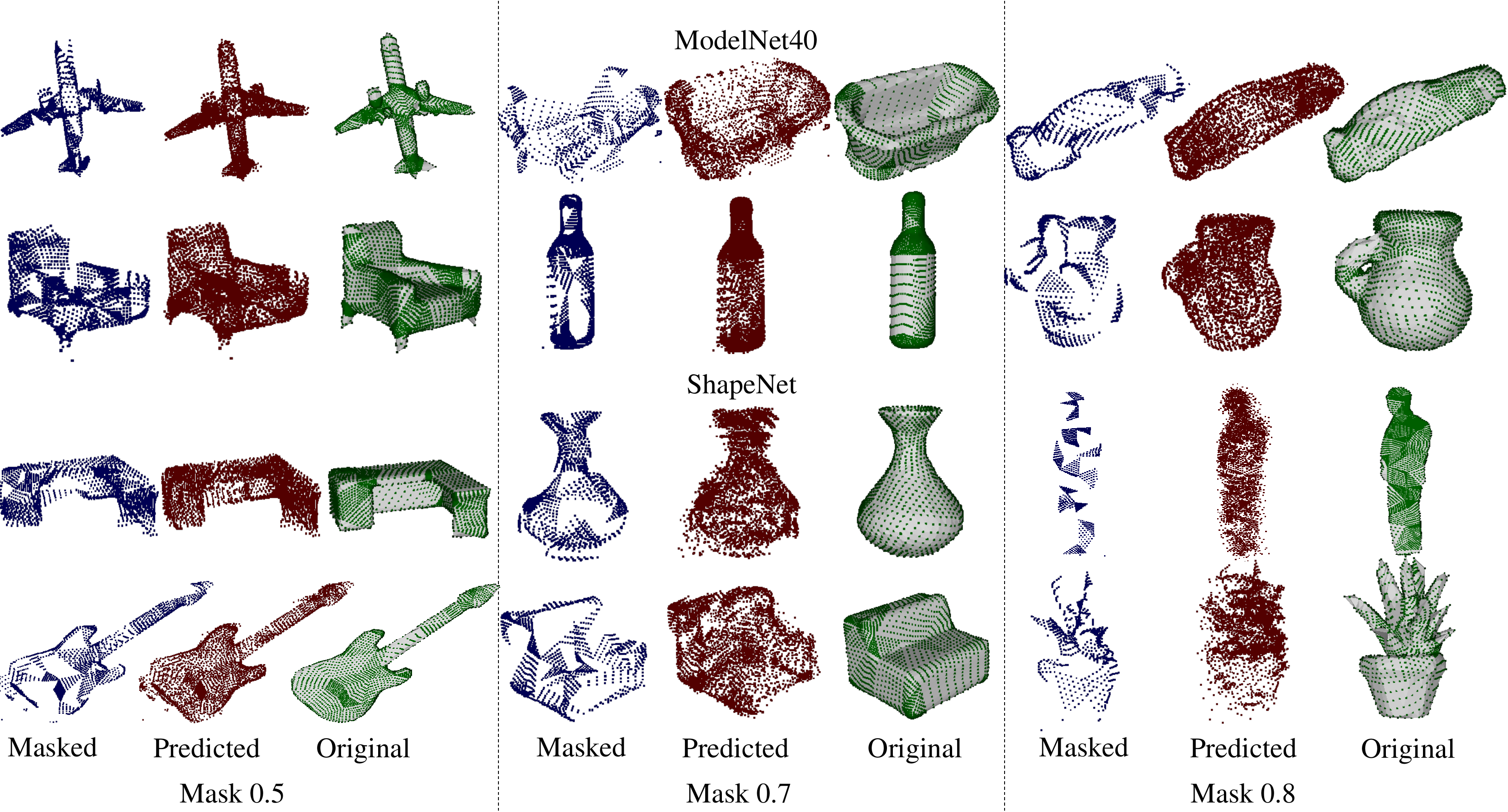}
    \caption{Reconstruction results on ModelNet40 test set. Top two lines are the results of model trained on ModelNet40, and bottom two lines are the results of model trained on ShapeNet. Here, we show the effects when the mask ratio is 0.5, 0.7 and 0.8 respectively.     More results will be demonstrated in the Supplementary.     }
    \label{fig:maskpa}
\end{figure*} 

It is noted that the input features do not contain coordinates of the three vertices of each face, while we propose to predict the vertices coordinates of the masked patches. Here, we illustrate the shape reconstruction effect on the ModelNet40 in Figure \ref{fig:maskpa}, where the  models are pre-trained on ModelNet40 and ShapeNet, respectively.
We directly show the vertices in mesh instead of faces, because the vertices are disordered and it is difficult to define the edge connection between them, which has also no impact on the pre-training task. It can be seen that our model can recover the shape of original meshes precisely by predicting vertices, and the performance of our model is still well even if the mask ratio is 80\%,  suggesting that the proposed model has indeed learned the geometric information of 3D meshes.

\section{Experiments}

Here, we first describe the data pre-processing methods and introduce the settings of the proposed pre-training scheme. The learned  representation would be evaluated  through doing supervised training under two settings: end-to-end fine-tuning and linear probing. The experiments are conducted on two downstream tasks, including object classification and part segmentation.  We provide more experimental results, analyses, and visualizations in the Supplementary.

\begin{table}[t]
\centering
\caption{Classification accuracy on  ModelNet40.  The first row is a point cloud method.}
\label{Classification}
\begin{tabular}{c|c}
\hline
Method & Acc (\%)\\ 
\hline
PCT~\cite{guo2021pct}  & 92.4\\
\hline
MeshWalker~\cite{lahav2020meshwalker}& 90.5\\
MeshNet~\cite{feng2019meshnet} & 88.4\\
SubdivNet ~\cite{hu2021subdivision}   & 91.4\\ \hline 
Transformer (w/o PT) & 91.5 \\
MeshMAE (PT M) & 91.7 \\
MeshMAE (PT S) & 92.5 \\
\hline
\end{tabular}
\end{table}

\subsection{Implementation Details}
\subsubsection{Data Pre-processing.}

In the original mesh datasets, \textit{e.g.}, ShapeNet and ModelNet40, the number of faces varies dramatically among meshes and most 3D meshes are not watertight or 2-manifold, that is, there are holes or boundaries on the surface.  In order to facilitate the subsequent mesh preprocessing, we first reconstruct the shapes to build the corresponding manifold meshes and simplify the manifold meshes into 500 faces uniformly. 
Then, we remesh all meshes in the datasets  using MAPS algorithm \cite{lee1998maps,liu2020neural} to obtain the meshes with regular structures. In this process, the input mesh (500 faces) would be simplified to a base mesh with a  smaller resolution (96-256 faces), and  the base mesh would be further  subdivided 3 times. After pre-processing, the remeshed meshes contain $n \times 64$ faces, where $n$ denotes the face number in the base mesh.  To improve the data diversity, we generate 10-times remeshed meshes for each input mesh by collapsing random edges in the remeshing process.

\subsubsection{Data Augmentation.}

We apply random anisotropic scaling with a normal distribution $\mu = 1$ and $\sigma =0.1$ to reduce network sensitivity to the size of meshes. According to the geometrical characteristics of 3D meshes, we additionally utilize shape deformation. The shape deformation is based on the free form deformation (FFD) \cite{sederberg1986free}, driven by moving the position of external lattice control points.

\subsubsection{Training Details.} 

For pre-training, we utilize ShapeNet and ModelNet40 as the dataset respectively to explore the  difference between small datasets and large datasets. Among them, ShapeNet covers about 51, 000 3D meshes from 55 common object categories and ModelNet40 contains 12, 311 3D meshes from 40 categories.  We utilize  ViT-Base \cite{dosovitskiy2020image}  as the encoder network with very slight modification, \textit{e.g.,} the number of input features' channels. And following \cite{he2022masked}, we set a lightweight decoder, which has 6 layers. We employ an AdamW optimizer, using an initial learning rate of 1e-4 with a cosine learning schedule. The weight decay is set as 0.05 and the batch size is set as 32. We set the same encoder network  that of pre-training in downstream tasks. For the classification task, we exploit the max-pooling operation behind the encoder and append a linear classifier. While for the segmentation task, we utilize two segmentation heads to provide a two-level feature aggregation, which would be introduced in the following part. 
We set the batch size as 32, and employ AdamW optimizer with an initial learning  rate of  1e-4. The learning rate is decayed by a factor of 0.1 at 30 and 60 epochs in classification (80 and 160 epochs in segmentation).

\begin{figure*}[ht]

    \centering
    \includegraphics[width=0.9\linewidth]{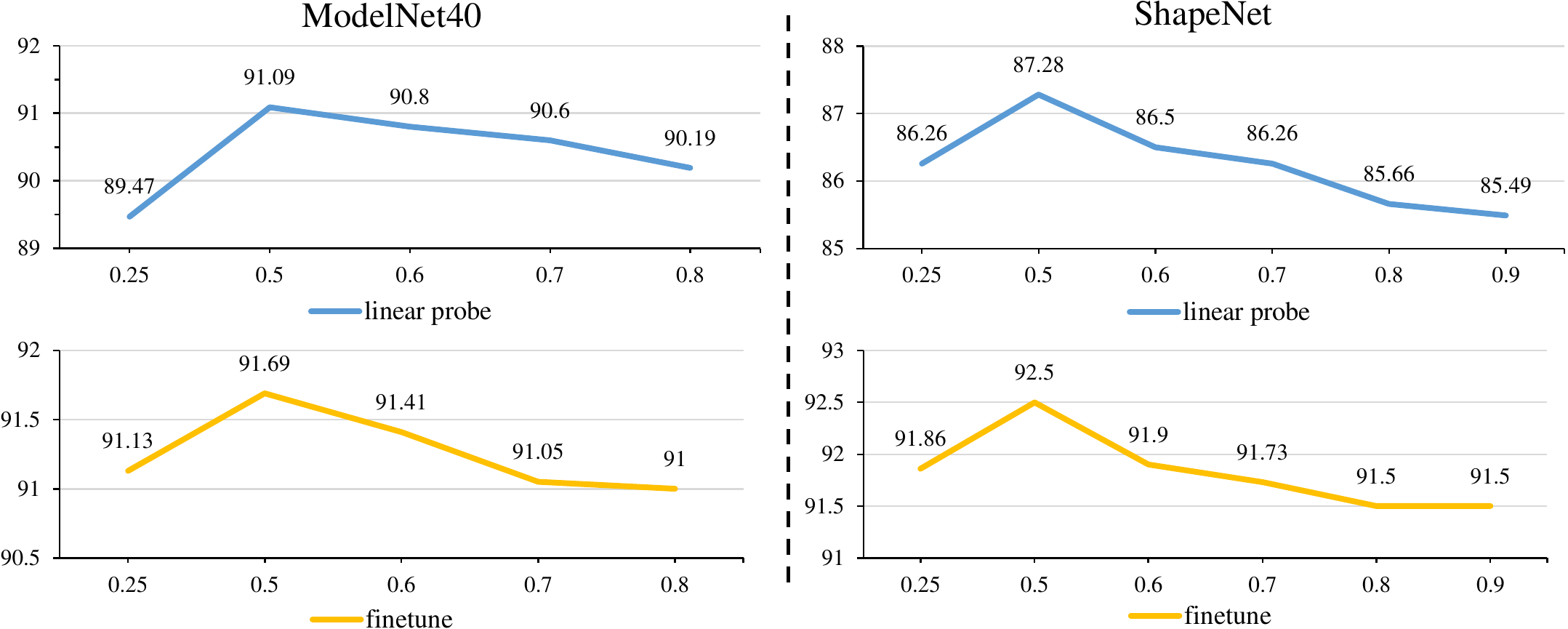}
    \caption{The experimental results under different masking ratios, where the pre-trained models are obtained on ModelNet40 in the left two figures while that are obtained on ShapeNet in the right two figures. The x-axes denotes the masking ratio and y-axes denotes the classification accuracy.}
    \label{fig:mask}
\end{figure*}

\subsection{Downstream Tasks}
\label{downstream}
In this subsection, we present the  experimental results of  classification and segmentation tasks. Here, we demonstrate the results of Mesh Transformer  without pre-training (`w/o PT'), the results of MeshMAE with pre-training on ModelNet (`PT M') and  ShapeNet (`PT S').

\subsubsection{Object Classification.}

For the classification task, we utilize the Manifold40~\cite{hu2021subdivision}  (generated from ModelNet40) as the dataset. To demonstrate the effectiveness of the proposed method, we compare it with several recent mesh classification methods and  a recent point cloud-based method. Experimental results are reported in Table~\ref{Classification}. In this part, we denote the results of  mesh Transformer  as `Transformer'. Here, we list the fine-tuning results of models pre-trained  on ModelNet40 and ShapeNet, respectively. The results of them both outperform the baseline, demonstrating that the proposed mask autoencoding strategy is also effective in the mesh analysis task. Compared with the result of `ModelNet40', the result of `ShapeNet' is higher and obtains the state-of-the-art performance, which shows that larger data sets can promote to learn a  better feature representation in the proposed pre-training task. This is  consistent with the conclusion in the CV.

Figure \ref{fig:mask} shows the linear probing and fine-tuning results under different masking ratios on ModelNet40. Though the number of meshes in ShapeNet is almost 5 times that in ModelNet40,  the linear probing results of ShapeNet are always lower than that of ModelNet40. We think this is due to the domain gaps between ShapeNet and ModelNet40. And the fine-tuning results of ShapeNet are much higher than that of ModelNet40, demonstrating a larger pre-training dataset could indeed bring better results. From Figure \ref{fig:mask}, we find that masking 50\% patches works well for both fine-tuning  and linear probing settings.

\subsubsection{Part Segmentation.} 

In this part, we utilize two datasets to conduct segmentation experiments, Human Body~\cite{maron2017convolutional} and COSEG-aliens~\cite{wang2012active}. Specifically, the Human Body dataset consists of  381 training meshes from SCAPE~\cite{anguelov2005scape}, FAUST~\cite{bogo2014faust}, MIT~\cite{vlasic2008articulated}, and Adobe Fuse~\cite{adobe2021}, and 18 test meshes from SHREC07~\cite{giorgi2007shape}. Each mesh sample is segmented into 8 parts. 
We also evaluate our method on the COSEG-aliens dataset with 200 meshes labeled with 4 parts.   Both two datasets are processed by the remeshing operation, and the face labels are obtained from the mapping between the remeshed data and the raw meshes using the nearest-face strategy. Compared with classification, the segmentation task is more challenging, since it needs to  predict dense labels for each face while faces within a patch might belong to different categories. In practice, we utilize two segmentation heads to provide a two-level feature aggregation. Specifically, we concatenate the output of the encoder with the feature embedding of each face to provide a fine-grained embedding. Besides, we do not design any other network structure specific for segmentation. Tables \ref{Coseg} and \ref{human} show the segmentation results on Human and COSEG-aliens datasets. Our method achieves comparable performance with recent state-of-the-art methods in segmentation tasks. Considering that the proposed method only uses patch embeddings, it might be difficult for the network to learn detailed structure information. Therefore, the performance improvement on the segmentation task is very limited compared with it on the classification task. Despite this, pre-training can still bring  improvements.

\begin{minipage}{\textwidth}
\begin{minipage}[t]{0.46\textwidth}
\flushleft
 \makeatletter\def\@captype{table}\makeatother\caption{Mesh segmentation accuracy on the COSEG-aliens dataset.}

\begin{tabular}{c|c}
\hline
Method& Acc (\%)\\
\hline
MeshCNN~\cite{hanocka2019meshcnn:} &94.4\\
PD-MeshNet~\cite{milano2020primal} & 89.0\\
SubdivNet~\cite{hu2021subdivision} & 98.5\\\hline
Transformer (w/o PT)&97.6\\
MeshMAE (PT M)&97.9\\ 
MeshMAE (PT S)&98.0\\\hline
\end{tabular}
\label{Coseg}
\end{minipage}
\begin{minipage}[t]{0.46\textwidth}
\centering
\makeatletter\def\@captype{table}\makeatother\caption{Mesh segmentation accuracy on the Human Body dataset.}

\begin{tabular}{c | c}
\hline
Method& Acc (\%) \\
\hline
Toric Cover~\cite{maron2017convolutional} & 88.0\\
MeshCNN~\cite{hanocka2019meshcnn:}& 87.7\\
SNGC~\cite{haim2019surface}& 91.3\\
SubdivNet~\cite{hu2021subdivision}& 90.8\\\hline
Transformer (w/o PT)& 90.1 \\
MeshMAE (PT M)& 90.1\\ 
MeshMAE (PT S)&90.3\\
\hline
\end{tabular}
\label{human}
\end{minipage}
\end{minipage}

\subsection{Ablation Studies}

Here, we present several ablation studies to further analyze the components in our method. To facilitate the analysis, we only conduct pre-training on ModelNet40 in this subsection.

\subsubsection{Reconstruction Target.}

In the proposed masked autoencoding task, we propose to recover the input feature and predict the vertices of masked patches simultaneously. In this part, we would verify their effectiveness separately and find an appropriate loss weight between them. In Table \ref{reconstruct}, we list the classification performance under several loss settings when the mask ratio is set as  0.5. It is found that the best classification performance can be obtained when $\mathcal{L}_{MSE} / \mathcal{L}_{CD} = 1:0.5$.

\begin{minipage}{\textwidth}
\begin{minipage}[t]{0.58\textwidth}
\flushleft
 \makeatletter\def\@captype{table}\makeatother\caption{Comparison of reconstruction targets. The values of $\mathcal{L}_{MSE}$ and  $\mathcal{L}_{CD}$ denote the weights of them, respectively. `Fine' indicates the results of fine-tuning, while `Line' indicates the results of linear probing.}
\label{reconstruct}
\begin{tabular}{c|c|c|c}
\hline
$\mathcal{L}_{MSE}$ & $\mathcal{L}_{CD}$  & Fine (\%) & Line (\%)\\
\hline
1&0&91.2&89.8\\
0&1&90.6&89.7\\ \hline
1&0.1&90.8&90.4\\
1&0.25&90.8&90.7\\
1&0.5&91.7&91.1\\
1&1&91.5&90.3\\
1&2&91.5&89.7\\
\hline
\end{tabular}
\end{minipage}
\begin{minipage}[t]{0.4\textwidth}
\centering
\makeatletter\def\@captype{figure}\makeatother\caption{The order of face features, where the serial number of the face denotes the location of its feature vector in the concatenated feature vector.}

\includegraphics[width=1\linewidth]{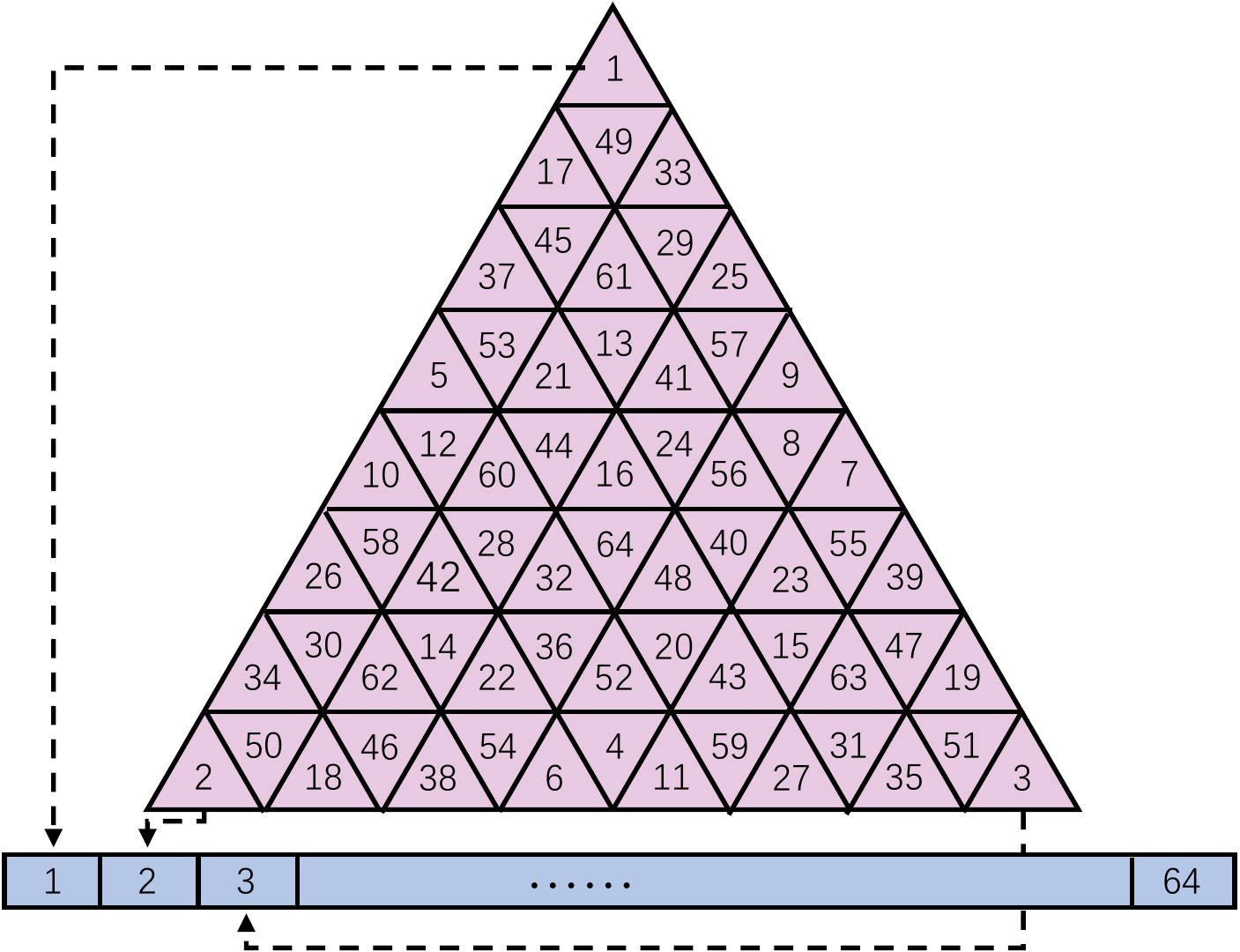}

\label{fig:order}

\end{minipage}
\end{minipage}

\subsubsection{Face Order.}
\label{faceorder}

In comparison with images with regular structure, the faces in patches are triangular and they cannot be arranged by row and column.  In this paper, we set the order of face features  in a patch as shown in Figure \ref{fig:order}, which is defined according to the remeshing process. 
Here, we conduct several test experiments to verify whether the other orders of faces would lower the performance. In Table \ref{order}, `Original' denotes the faces order in our method, `Rotate’ denotes the structure of faces remains unchanged while starting from another two angles, \textit{e.g.}, starting from face 2 (Rotate-l) or face 3 (Rotate-r), and `Random' denotes the face features are arranged randomly. In Table \ref{order}, we repeat the experiments  5 times for each setting. According to the results, we find that even if the order of faces is randomly disturbed, the performance decreases very slightly and remains basically stable, which demonstrates the our selection of the order works well for patch representation.

\begin{minipage}{\textwidth}
\begin{minipage}[t]{0.46\textwidth}
\flushleft
\makeatletter\def\@captype{table}\makeatother\caption{The effect of different orders of face features in the patches.}
\label{order}
\begin{tabular}{c|c|c} 
\hline  
Sorting Method& Acc (\%) & Std.  \\\hline
 Original & 91.7 & 4e-2 \\
 Rotate-l & 91.7 & 5e-2 \\
 Rotate-r & 91.6 & 2e-2 \\
 Random & 91.6 & 4e-2 \\
 \hline 
\end{tabular} 

\end{minipage}
\begin{minipage}[t]{0.46\textwidth}
\centering
 \makeatletter\def\@captype{table}\makeatother\caption{Comparison of different positional embedding strategies.}
\label{posemb}
\begin{tabular}{c|c|c|c|c}
\hline	
Strategy& Mask& Acc (\%)& Mask& Acc (\%)\\
\hline
a) &0.25&42.6&0.5&46.1\\
b) &0.25&88.2&0.5&89.3\\
c) &0.25&75.0&0.5&77.5\\
d) &0.25&83.7&0.5&89.7\\
\hline
\end{tabular} 
\end{minipage}
\end{minipage}

\subsubsection{Positional Embedding.}

Each patch contains 64 faces, and how to embed positional information of 64 faces is a problem worth exploring. 
Here, we propose several positional embedding strategies: a) utilizing learnable parameters as MAE \cite{he2022masked} does;  b) first embedding the center coordinates of each face, then applying max-pooling; c) first reshaping the center coordinates of 64 faces (64, 3) into a one-dimension vector (64$\times$3), then embedding this vector directly; d) first calculating the center coordinates of the whole patches, then embedding the center of the patch directly (ours).  Table \ref{posemb} lists the classification results of the above strategies, which are all obtained by linear probing. We find that embedding the center of the patch directly could obtain the best results.

\section{Conclusions}

In this paper, we introduce Transformer-based architecture (Mesh Transformer) into mesh analysis and investigate the masked-autoencoding-based pre-training method for Mesh Transformer pre-training. By recovering the features and shape of the masked patches, the model could learn the geometric information of 3D meshes. And the comprehensive experimental results and ablations prove that the proposed MeshMAE could learn a more effective feature representation.  The study shows the feasibility of the Transformer-based method on mesh analysis. In future, we would like to explore its applications in other mesh-related tasks.

\textbf{Acknowledgements:}
This work is supported by the National Natural Science Foundation of China under Grant No. 62072348. Dr. Baosheng Yu and Dr. Jing Zhang are supported by ARC Project FL-170100117.

\bibliographystyle{splncs04}
\bibliography{maintext}

\end{document}